\documentclass[conference]{IEEEtran}
\IEEEoverridecommandlockouts
\usepackage{cite}
\usepackage{amsmath,amssymb,amsfonts}
\usepackage{algorithmic}
\usepackage{graphicx}
\usepackage{textcomp}
\usepackage{xcolor}
\usepackage{algorithm}
\usepackage{cleveref}
\usepackage{array} 
\usepackage{amsmath}
\usepackage{booktabs}

\def\BibTeX{{\rm B\kern-.05em{\sc i\kern-.025em b}\kern-.08em
    T\kern-.1667em\lower.7ex\hbox{E}\kern-.125emX}}
\begin{document}

\title{Control-CLIP: Decoupling Category and Style Guidance in CLIP for Specific-Domain Generation}

\author{\IEEEauthorblockN{
Zexi Jia\IEEEauthorrefmark{2}, 
Chuanwei Huang\IEEEauthorrefmark{2},
Hongyan Fei, 
Yeshuang Zhu, 
Zhiqiang Yuan, 
Jinchao Zhang*, 
Jie Zhou* }
\IEEEauthorblockA{Peking University, Beijing, China \\  Wechat AI, Tencent, China}
}

\maketitle

\begin{abstract}
Text-to-image diffusion models have shown remarkable capabilities of generating high-quality images closely aligned with textual inputs. However, the effectiveness of text guidance heavily relies on the CLIP text encoder, which is trained to pay more attention to general content but struggles to capture semantics in specific domains like styles. 
As a result, generation models tend to fail on prompts like "a photo of a cat in Pokemon style" in terms of simply producing images depicting "a photo of a cat".
To fill this gap, we propose Control-CLIP, a novel decoupled CLIP fine-tuning framework that enables the CLIP model to learn the meaning of category and style in a complement manner. With specially designed fine-tuning tasks on minimal data and a modified cross-attention mechanism, Control-CLIP can precisely guide the diffusion model to a specific domain. Moreover, the parameters of the diffusion model remain unchanged at all, preserving the original generation performance and diversity. Experiments across multiple domains confirm the effectiveness of our approach, particularly highlighting its robust plug-and-play capability in generating content with various specific styles.
\end{abstract}


\section{Introduction}
\label{sec:intro}

\begin{figure*}[ht]
\begin{center}
\includegraphics[width=0.9\linewidth]{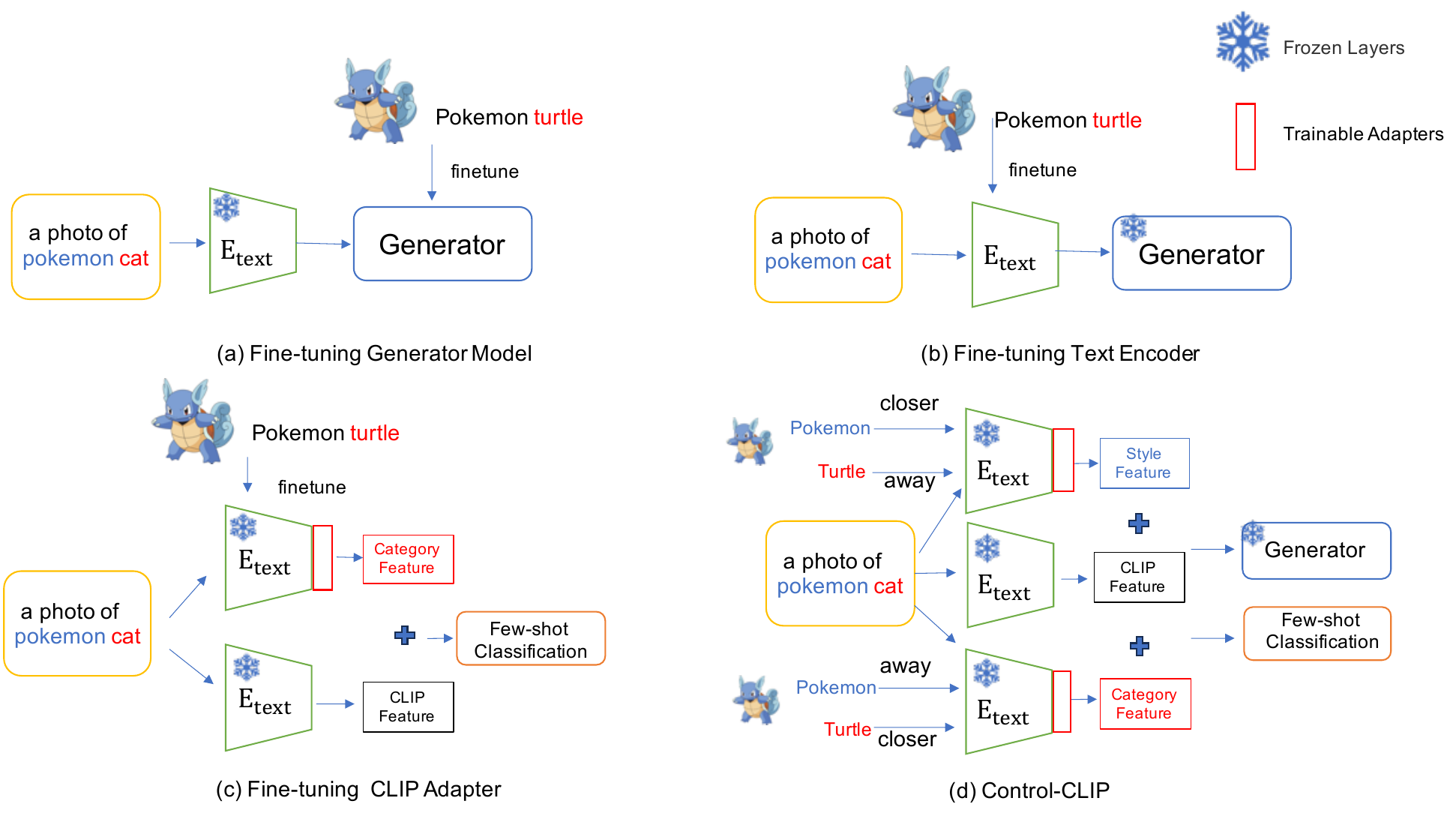} 
\end{center}
   \caption{Comparison of different approaches to leveraging in-domain text and image data for fine-tuning. Previously common fine-tuning methods include: (a) fine-tuning only the generator, (b) fine-tuning only the text encoder, and (c) using an adapter to decouple category features from the text encoder and combine them with the original text features. (d) Our Control-CLIP utilizes a decoupled design to learn both category and style features for generation models.}
\label{fig:first}
\end{figure*}

In recent years, there have been significant advancements in text-controlled image generation. With the emergence of text-to-image models~\cite{rombach2022high, jia2024finger,huang2025semantic,jia2023fingerstr,wang2023improving}, we can now create visually stunning images by providing textual prompts. 
Apart from generating high quality images, accurately controlling and adapting the style of generated images via text is also a crucial requirement in AI-based artistic creation. A challenge lies in leveraging domain-specific text-image data to help pre-trained generative models better understand the stylistic nuances of a particular domain and accurately generate images based on textual descriptions.
To tackle this problem, methods have been proposed to fine-tune the diffusion generation model~\cite{zhai2023investigating} or incorporating various control conditions~\cite{tumanyan2023plug}, as illustrated in Fig.~\ref{fig:first} (a) and (b).

Contrastive Language-Image Pre-training (CLIP)~\cite{radford2021learning} effectively bridges the gap between the language and image domains and serves as the foundational network for most diffusion-based image generation models.

However, despite the growing use of CLIP models in generative tasks like text-to-image and text-to-video generation, the need for domain adaptability in CLIP models and its impact on generative models has been under-explored. We propose that domain adaptation of text-to-image diffusion models can be effectively achieved by fine-tuning only the CLIP component on a small amount of in-domain data, thus preserving the general-domain generation capability of pre-trained diffusion models.

In this paper, we consider style features as domain-specific and object categories as domain-invariant. As noted in domain adaptation~\cite{wang2018deep}, each domain has unique characteristics, while all domains share invariant features. For instance, we can differentiate images like cartoons, sketches, and real photos based on color strokes and texture, which are domain-specific features. However, all styles of images can depict the same object, such as a cat or a dog, determined by category features independent of the domain.

Previous research on fine-tuning the CLIP model for domain adaptation mainly focused on learning category features to improve classification accuracy within specific domains, as shown in Fig.~\ref{fig:first}(c). However, these methods do not explicitly separate style features from category features in their loss functions. While effective in some cases, they inadvertently reduce the model's ability to recognize stylistic nuances. In the application of text-to-image generation, a depth understanding of nuanced domain-specific styles is demanded. Traditional fine-tuning methods of CLIP models focus on learning domain-invariant features for classification has inadvertently suppressed the model's sensitivity to style-specific attributes. This tendency has impeded the fine-tuned CLIP models in recognizing various stylistic forms and has constrained their ability to accurately interpret text descriptions about domain-specific styles.

This paper introduces a novel approach called \emph{Control-CLIP}, designed to explicitly disentangle style and category features to enhance the representation capacity of CLIP. Our framework uses two distinct text encoders—style and category encoders—fine-tuned independently. As shown in Fig.~\ref{fig:first}(d), this integration improves the CLIP model's ability to interpret textual descriptions of object categories and styles. By decoupling domain features, the CLIP model gains enhanced resilience in style and category perception, excelling in precise few-shot classification and style discrimination tasks within specific domains. This improved semantic understanding also allows the CLIP model to better control image generation, producing images that align more accurately with language descriptions. We develop two variations of the Control-CLIP model to cater to diverse datasets. One variant uses cross-entropy-based loss functions for datasets with style labels, while the other employs triplet loss for datasets with only text description labels. This adaptability demonstrates the method's efficacy in enhancing the CLIP model's comprehension of both style and category information.

Additionally, we introduce a novel cross-attention mechanism that combines the fine-tuned outcomes of the style and category encoders with the Stable Diffusion~\cite{rombach2022high} model. This refined structure integrates seamlessly with Stable Diffusion in a plug-and-play manner, eliminating the need to train the generative model. Incorporating the Control-CLIP model enhances the generative model's understanding of textual style and category descriptions, improving the generation of in-domain images that closely match the provided descriptions.

In summary, the main contributions of this paper include:
\begin{itemize}
\item We propose Control-CLIP, a CLIP based text-image alignment model that explicitly decouples style and category features.
\item  Control-CLIP can be fine-tuned on specific domain datasets, and it can enhance the category discrimination ability and style discrimination ability of the CLIP model simultaneously in specific domains. 
\item We propose a new form of cross-attention that combines Control-CLIP into the Stable Diffusion model. This integration helps generate images that align with semantic descriptions without adjusting the generative model.
\end{itemize}
\section{Related work}

\textbf{Domain Adaptation CLIPs:} With the emergence of the pre-trained image-text model CLIP, adapting image features more accurately has become possible. Current research focuses on leveraging CLIP-extracted features to enhance few-shot learning. Methods like CoOp~\cite{zhou2022learning} and CoCoOp~\cite{zhou2022conditional} use learnable vectors to simulate contextual vocabulary for prompts while keeping pre-trained parameters unchanged. CLIP-Adapter~\cite{gao2023clip} fine-tunes with adapters on visual or language branches. Tip-Adapter~\cite{zhang2021tip} create weights from a key-value cache model from a few-shot training set without backpropagation. TPT~\cite{shu2022test} uses consistency among multiple views of the same image as a supervision signal for prediction. These methods help CLIP learn domain-invariant information for classification but reduce the ability to learn domain-specific information. Our paper aims to enrich CLIP's understanding of image styles and categories and explore better assistance for downstream generative tasks.

\textbf{Fine-tuning Large Generative Models: }
Fine-tuning large-scale generative models on specific tasks aims to adapt the model to specific datasets or tasks, enhancing performance in specialized applications. \textbf{Finetuning}~\cite{kumar2022fine} involves continuing training with additional data, but risks overfitting, mode collapse, and catastrophic forgetting. Extensive research has focused on strategies to mitigate these issues. \textbf{Adapter}~\cite{chen2022vision} technology adds extra "adapter" layers to pre-trained models, allowing better adaptation to downstream tasks with fewer parameters, reducing computational resources and storage costs. \textbf{Prefix-Tuning}~\cite{li2021prefix} adjusts model behavior by adding continuous prefix vectors before the input sequence. \textbf{Low-Rank Adaptation (LoRA)}~\cite{hu2022lora} reduces computational resources and storage costs by performing parameter updates in a low-rank subspace while maintaining performance. However, fine-tuning large models reduces recognition accuracy for generic object categories and affects generated content diversity.

\section{Method}

\subsection{Decoupling Style and Category Features}

\begin{figure*}[h!]
\begin{center}
\includegraphics[width=0.9
\linewidth]{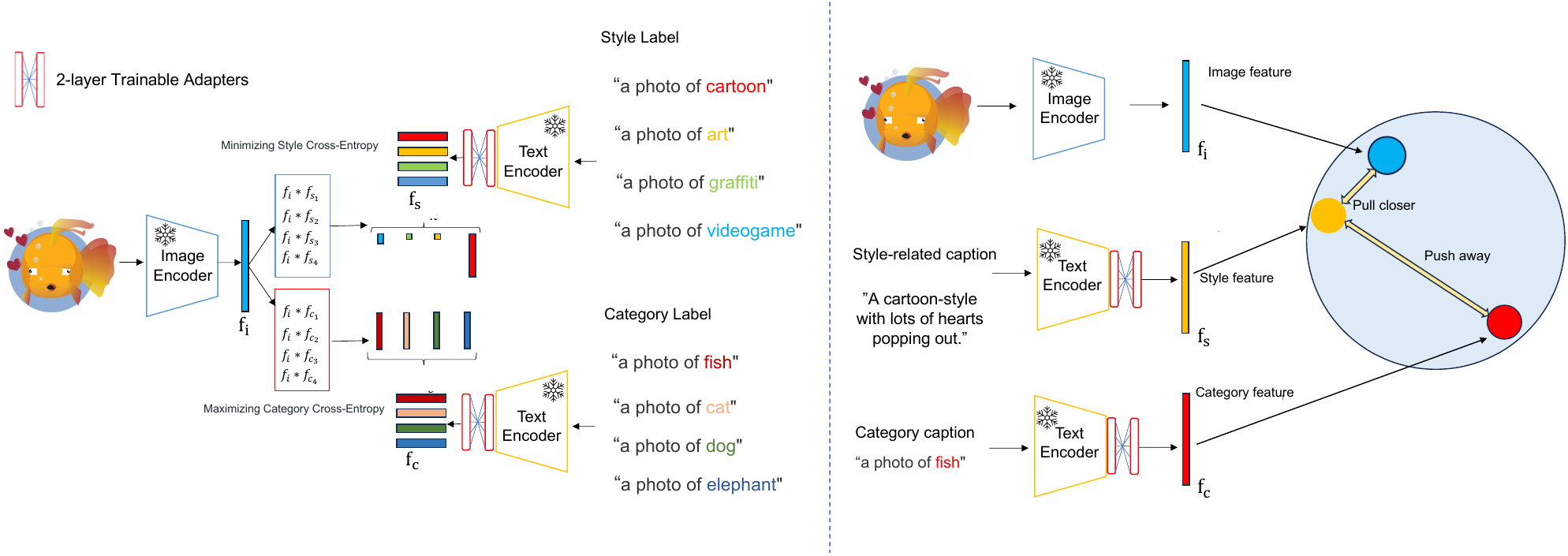}
\end{center}
   \caption{Training style encoder in Control-CLIP. Left: On datasets with style labels, Control-CLIP separates category and style features through cross-entropy loss, achieving better performance but requires higher annotation requirements.  Right: On datasets without style labels, we can utilize image captions to obtain style information. Control-CLIP uses a triplet loss function to distinguish between category and style features. The category encoder is trained similarly with an inverted loss function.
}
\label{fig:train}
\end{figure*}

Inspired by the work of Adversarial Representation Learning (ARL)~\cite{ding2022domain}, our approach involves a dual strategy of minimizing cross-entropy associated with target labels and maximizing cross-entropy for non-target labels. This method aims to separate domain-specific features and domain-invariant features effectively. To achieve this, we utilize two distinct CLIP-based fine-tuning networks, referred to as the style encoder and the category encoder, respectively. We use $f$ to represent the output of CLIP, and $f_{s}$ to represent the output of the style encoder, $f_{c}$ to represent the output of the category encoder, and $f_i$ to represent the output of image encoder. The primary objective is to endow each fine-tuned model with discriminative capabilities pertinent to its designated domain. Specifically, the style encoder is fine-tuned to minimize style classification errors while intentionally maximizing category classification errors. This approach is designed to encourage predictions for categories to approximate a uniform distribution. Conversely, the category encoder is fine-tuned to exhibit the opposite behavior, prioritizing accuracy in category classification and less focus on style differentiation. The training pipeline is illustrated in Fig.~\ref{fig:train}.

\textbf{Labelled Data}: When applying fine-tuning on datasets with style labels, we leverage two cross-entropy loss functions to effectively distinguish between style and category attributes. The style encoder's objective is to minimize style-related loss while concurrently maximizing category-related loss. The formulation of this approach is as follows:
\begin{equation}
\begin{aligned}
&\mathcal{L}_{1}=-\sum \operatorname{softmax}(f_{s}) \log y_{s} \\
&\mathcal{L}_{2}=\sum \operatorname{softmax}(f_{s}) \log {y_{c}},
\end{aligned}
\end{equation}
where ${y_s}$ and $y_c$ represent the style label and category label of the image, respectively. During the training process, we simultaneously constrain both of them, resulting in the total loss function:
\begin{equation}
\mathcal{L}_{style}^{label}=\mathcal{L}_{1}+ {\lambda}_{1} \mathcal{L}_{2}.
\end{equation}
Correspondingly, the loss function for the category encoder during fine-tuning is as follows:
\begin{equation}
\begin{aligned}
&\mathcal{L}_{1}^{'}=-\sum \operatorname{softmax}(f_{c}) \log y_{c} \\
&\mathcal{L}_{2}^{'}=\sum \operatorname{softmax}(f_{c}) \log y_{s},
\end{aligned}
\end{equation}
and the total loss function for the category encoder is:
\begin{equation}
\mathcal{L}_{category}^{label}=\mathcal{L}_{1}^{'}+ {\lambda}_{2} \mathcal{L}_{2}^{'}.
\end{equation}


\textbf{Unlabelled Data}: Most fine-tuning datasets are comprised of image-text pairs without explicit style labels. 
Generally, in generative scenarios, we can obtain captions corresponding to images. For images without captions, Vision-Language Models (VLM) can be utilized to generate them. We apply a strategy where we filter out category-related words using regular expression techniques, leaving behind primarily style-related text. The style-related text is then inputted into the text encoder of the CLIP model, extracting the \textbf{style feature} $f_{s}$, and feeding the category-related text to obtain the \textbf{category feature} $f_{c}$. 
Due to the absence of corresponding style labels, we utilize $f_i$ as an anchor to ensure the retention of style information in $f_s$. Simultaneously, by increasing the distance between $f_s$ and $f_c$, we aim to eliminate category information from $f_s$. More specifically, we employ triplet loss~\cite{schroff2015facenet} to implement the aforementioned constraints for the style encoder:
\begin{equation}
\mathcal{L}_{style}^{unlabel}= \sum \max (\|f_{s}-f_{i}\|_2 - \|f_{s}-f_{c}\|_2+m_1, 0).
\end{equation}
Correspondingly, the loss function for category encoder is:
\begin{equation}
\mathcal{L}_{category}^{unlabel}= \sum \max (\|f_{c}-f_{i}\|_2 - \|f_{c}-f_{s}\|_2+m_2, 0).
\end{equation}
$m_1$ and $m_2$ represent the margins in the triplet loss. In this paper, we set $m_1 = m_2 = 0.3$.

In our model, we introduce an Adapter module consisting of two learnable linear layers with activation functions. This compact module efficiently captures domain-specific style and category characteristics, streamlining the learning process and reducing overfitting with minimal parameters.

For few-shot classification, we combine outputs from the category encoder and CLIP's text encoder to obtain finely tuned category-specific features. In the image style discrimination task, we use a weighted sum of the style encoder and CLIP's text encoder outputs. For generation tasks, we guide the process using both style and category features in a plug-and-play manner within the diffusion model.

\subsection{Using Control-CLIP as Generation Condition}
The research on Stable Diffusion~\cite{rombach2022high} demonstrates that diffusion models can model conditional distributions of the form $p(z|y)$. This can be achieved through the use of a conditional denoising autoencoder $\theta(z_t, t, y)$. 
\begin{figure}[t]
\begin{center}
\includegraphics[width=0.9\linewidth]{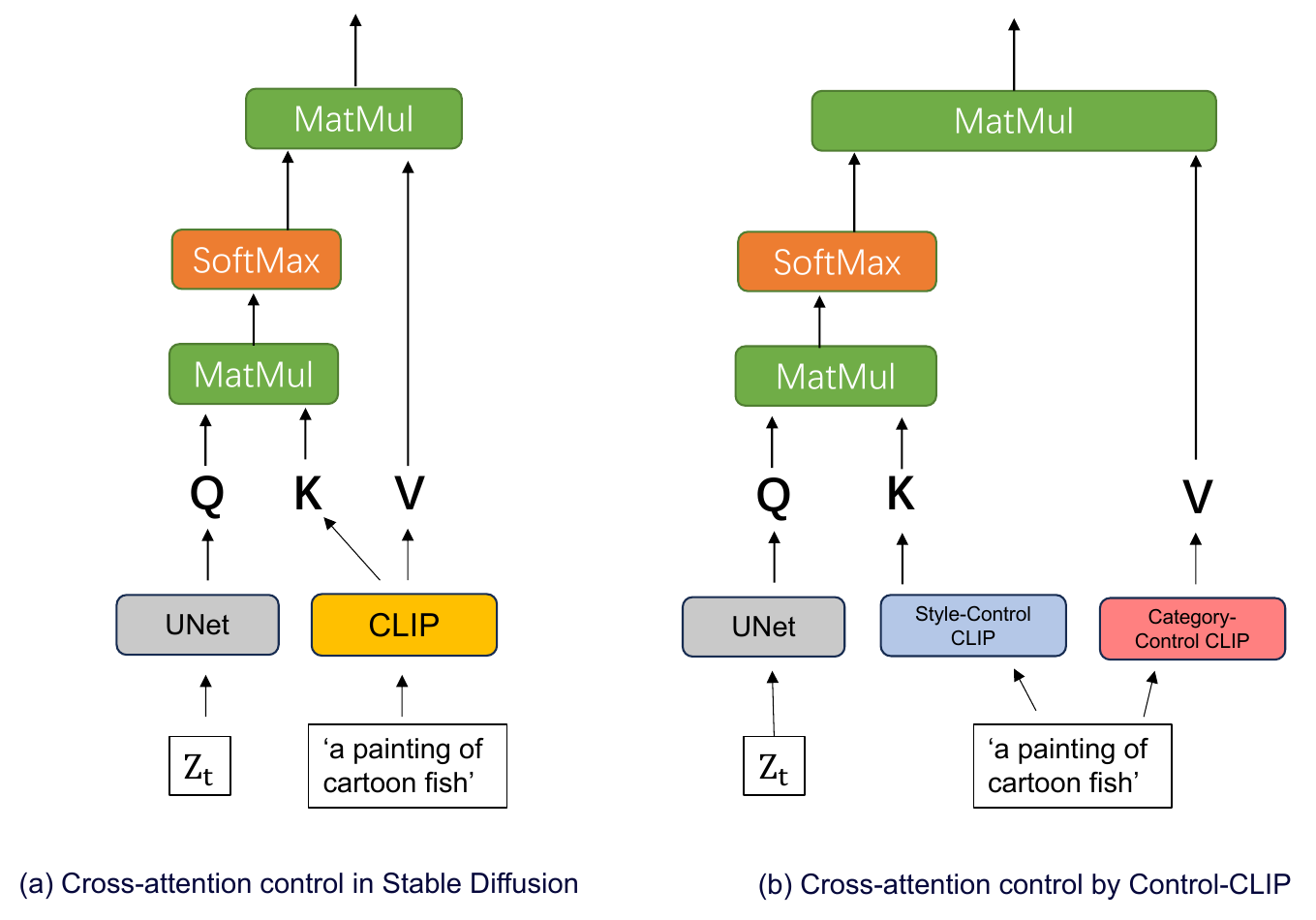}
\end{center}
   \caption{We replace the input of the K and V matrices in the attention mechanism with the outputs of Style-Control and Category-Control CLIP.}
\label{fig:free}
\end{figure}
In this paper, a new control method is proposed, where we expand $\theta(z_t, t, y)$ to $\theta(z_t, t,f_s,f_c)$, allowing for joint control of image generation through decoupled style and category features, as shown in Fig.~\ref{fig:free}. 
We draw inspiration from Stable Diffusion and use a cross-attention mechanism to integrate style and category control.
Regarding the original attention mechanism in Transformer:
\begin{equation}
\operatorname{Attention}(Q, K, V)=\operatorname{softmax}\left(\frac{Q K^T}{\sqrt{d}}\right) \cdot V,
\end{equation}
where $Q$,$K$,$V$originate from the same input.
However, in Stable Diffusion, $Q$ is derived from the noise prediction network $\varphi_i\left(z\right)$, while $K$ and $V$ originate from the same control condition encoder $\tau_\theta(y)$:
\begin{equation}
Q=W_Q^{(i)} \cdot \varphi_i\left(z\right), K=W_K^{(i)} \cdot \tau_\theta(y), V=W_V^{(i)} \cdot \tau_\theta(y).
\end{equation}

In this paper, we distinguish the sources of the K and V matrices. Specifically, the K matrix is derived from the style encoder $\tau_{style}$, whereas the V matrix is obtained from the category encoder $\tau_{category}$ , as illustrated in the formula:

\begin{equation}
K=W_K^{(i)} \cdot \tau_{style}(y), V=W_V^{(i)} \cdot \tau_{category}(y).
\end{equation}
By default, we employ style encoder $\tau_{style}$ and category encoder $\tau_{category}$ to control the generation process together. 
The extracted $f_s$ and $f_c$ are aggregated with the original CLIP text feature $f_{text}$ through residual connections:

\begin{equation}
    \begin{aligned}
    &\tau_{style}(y) = \alpha f_{s} + (1- \alpha) f_{text},\\
    &\tau_{category}(y) = \alpha f_{c} + (1- \alpha) f_{text}.
    \end{aligned}
\label{eq:alpha}
\end{equation}
For generation task, the residual ratio $\alpha$ is set to 0.1.


\section{Experiments}

\begin{table*}[]
\begin{center}
\caption{The category and style classification accuracy of CLIP and CLIP-based fine-tuned models on 4 datasets. The comparative work is based on training with style labels.}
\label{table:more}
\setlength{\tabcolsep}{3.5mm}{
\begin{tabular}{lcccccccc}
\toprule
Top-1 Accuracy  & \multicolumn{2}{c}{ImageNet-R} & \multicolumn{2}{c}{PACS} & \multicolumn{2}{c}{VLCS} & \multicolumn{2}{c}{OfficeHome} \\
\midrule
    & Style       & Category  & Style       & Category       & Style   & Category      & Style     & Category   \\
    \midrule
CLIP~\cite{radford2021learning}  &56.0 &46.5  &  69.8            & 88.3             &    25.0       &  69.8           &       50.3    & 71.8             \\
CoOp~\cite{zhou2022learning}    &57.8 &66.4     &   82.1      &  92.3              &   90.0        &73.6         &        86.7   & 74.0            \\
CoCoOp~\cite{zhou2022conditional} &57.5 &66.2  &      80.5        &   \textbf{93.7}        &    88.2       &  74.8           &   87.1        & 73.5             \\
CLIP-Adapter~\cite{gao2023clip} &63.1 &65.3   &      79.8        & 91.2              &     85.6      & 75.1           &        85.5   & 74.3 \\
TPT~\cite{shu2022test}   &59.1 &50.2   &       69.8       &      90.9           &      25.0     &  74.8         &    50.3       & 73.6 \\   
\midrule
Control-CLIP  &\textbf{77.4} &\textbf{70.5} &          \textbf{83.4}    &    92.4            &  \textbf{90.2}        &  \textbf{75.9}            &     \textbf{87.7}      & \textbf{74.7}  \\   
Control-CLIP(caption)  &76.9 &64.2 &       78.9       &  91.7               &   83.1        & 72.4             &      81.2     &72.9  \\   
\bottomrule
\end{tabular}
}
\end{center}
\end{table*}

\subsection{Experimental Settings}
\textbf{Datasets:} We conduct experiments with Control-CLIP on four image classification datasets with well-defined styles and category labels, namely ImageNet-R~\cite{hendrycks2021many}, PACS~\cite{li2017deeper}, VLCS~\cite{li2017deeper}, and OfficeHome~\cite{venkateswara2017deep}. These data include two different labels, one representing categories and the other representing styles. For example, in ImageNet-R, there are 200 object category labels that are the same as those in ImageNet, as well as 16 different style labels such as cartoon, video game, painting, and so on.

\textbf{Training:} All methods are trained with 16 samples and evaluated on the full test set. We use manually crafted hard prompts for text input, following CLIP's approach. For datasets with both style and semantic categories, like ImageNet-R, we validate our style-based and description-based models separately. The style-based model uses the prompt ``a photo of {STYLE} {Category}," while the description-based model uses randomly selected common image description templates. We use CLIP's ResNet-50 and VIT-B/16 as visual encoders and a Transformer as the text encoder, keeping pre-trained weights fixed. Data pre-processing follows CLIP's protocols, including random cropping, resizing, and horizontal flipping..
\subsection{Performance on Discrimination}
We first evaluate our method on the discriminative task of category and style recognition. We fine-tuned the model on a 16-shots setting of ImageNet-R with 30 epochs and tested its performance on the remaining data. As Tab.~\ref{table:more} shows, our method simultaneously improves the accuracy of category recognition and style recognition, achieving the highest accuracy in style classification. For a fair comparison, we also use category prompts and style prompts respectively for other CLIP-based fine-tuning methods. It is important to note that cache-based models, such as TIP~\cite{zhang2021tip} and APE~\cite{zhu2023not}, struggle to cache multiple attributes of images at the same time, and therefore cannot perform both category and style recognition simultaneously. Compared to full fine-tuning of CLIP, our method only needs to adjust the parameters of two fully connected layers, significantly reducing the number of parameters during training.

Tab.~\ref{table:more} also shows the results of style and category classification on more datasets. We used a 16-shot training set and trained the CLIP fine-tuning methods with style prompts with 30 epochs for fair comparison. The experimental results show that Control-CLIP achieves higher accuracy scores on both style and category across several datasets.

\begin{table}[]
\caption{Recognition accuracy with varying residual ratio $\alpha$.}
\setlength{\tabcolsep}{3.2mm}{
\begin{tabular}{lcccccc}
\toprule
Ratio $\alpha$                  & 0 & 0.2 & 0.4 & 0.6 & 0.8 & 1 \\
\midrule
Category & 56.0  & 62.5    &  \textbf{63.9}   & 63.9    & 63.5   & 63.2  \\
Style   & 46.5  &  66.3   &   70.3  &  70.5   & \textbf{71.4}   & 71.3 \\
\midrule
Ratio $\lambda$                  & 0 & 0.1 & 0.2 & 0.3 & 0.4 & 0.5 \\
\midrule
Category & 63.1  &  63.5   & \textbf{63.9}    &  63.7   & 63.0    & 62.4  \\
Style    & 65.3  &  66.1   &  67.9   &  \textbf{68.1}   & 67.4    & 63.1 \\
\bottomrule

\end{tabular}}
\label{table:alpha}
\end{table}

\subsection{Performance on Generation}

\begin{figure*}[t]
\begin{center}
\includegraphics[width=1.0\linewidth]{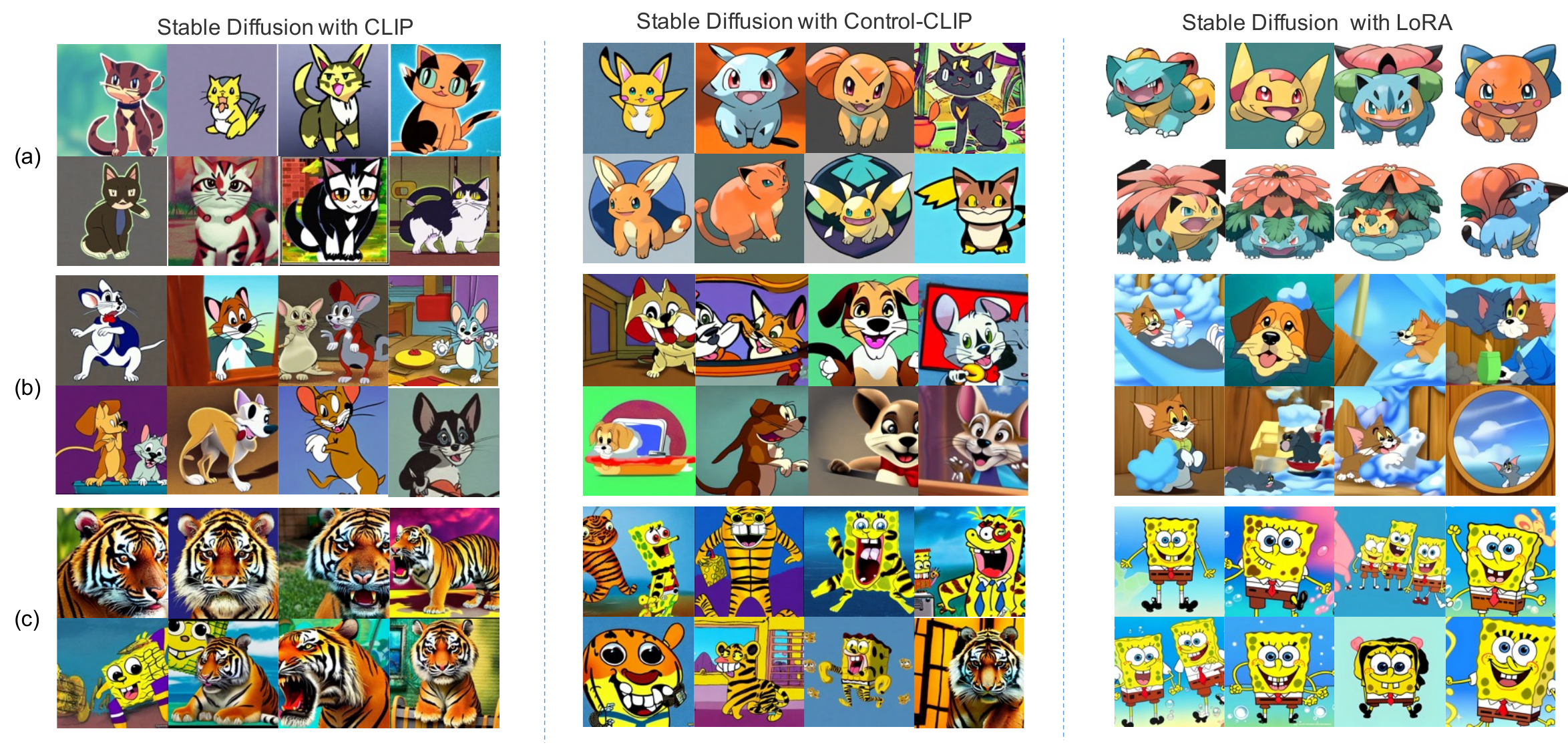}
\end{center}
   \caption{With the guidance of Control-CLIP, Stable Diffusion can generate images closely aligned with the prompts while maintaining the diversity of the generated contents. Compared to LoRA, Control-CLIP's generated results have better style and content diversity. Prompts: (a) \textcolor{blue}{A photo of Pokemon style cat;} (b) \textcolor{brown}{A photo of Tom and Jerry style dog;} (c) \textcolor{orange}{A photo of SpongeBob style tiger.} For each cartoon domain, we use the same set for LoRA and Control-CLIP of 10 images for fine-tuning.}
\label{fig:show}
\end{figure*}

We employed a multi-style cartoon dataset~\cite{kaggle} to evaluate the generation of diverse styles. This dataset comprises cartoon images in 10 distinct styles sourced from a range of classic animated series, such as Pokemon, Smurfs, South Park, SpongeBob SquarePants, Tom and Jerry, etc. From each style class, we randomly chose 10 images and labeled their corresponding category information to train the Control-CLIP model. After training, we integrated Control-CLIP into the Stable Diffusion v1.5 to test the efficacy of our model in guiding the generative process.

Fig.~\ref{fig:show} 
shows generation results under three cartoon-style datasets. These generation results are all generated from the same prompt template ``a photo of \{STYLE\} style \{Category\}", for example, ``a photo of Pokemon style cat". Control-CLIP demonstrates more accurate content and semantic understanding capabilities, resulting in more accurate generation results. 

As our work involved fine-tuning CLIP and enhancing its style understanding capabilities, objective metrics like FID and CLIP score are challenging to evaluate our generative work. Therefore, we use Average Human Ranking (AHR)~\cite{zhang2023adding} as a preference metric, where users rate each result on a scale from 1 to 5 (the higher the score, the better the generated quality). We invited 10 volunteers, who did not participate in this work, to score the generative quality and description fit of 10 results for each of 10 cartoon categories. The Control-CLIP guidance helped Stable-Diffusion's generative quality average score increase from 1.84 to 2.70, and the description fit score from 2.52 to 3.78.

\subsection{Ablation Study}

\textbf{Residual ratio $\alpha$: } We study the residual ratio $\alpha$ in Eq.~\ref{eq:alpha}. Tab.~\ref{table:alpha} shows that on ImageNet-R, the optimal $\alpha$ is 0.8 for style recognition and 0.4 for category recognition, indicating style adaptation needs more new knowledge. When $\alpha=0$, Control-CLIP becomes the original CLIP, and performance drops, showing successful decoupling of style and category features. At $\alpha=1.0$, reliance on the new CLIP model leads to overfitting.

\textbf{LOSS ratio $\lambda$: } We study the loss weight $\lambda$ to assess our adversarial training strategy. Tab.~\ref{table:alpha} shows that adversarial training improves recognition accuracy for both categories and styles. When $\lambda=0$, Control-CLIP becomes CLIP-Adapter, causing a performance drop.
\section{Conclusion}
This paper introduces Control-CLIP, a novel framework that enhances its ability to accurately understand concepts in specific domains by decoupling style from category features. We establish independent style and category control modules and apply different loss functions to fine-tune the CLIP model. Compared to other fine-tuning methods, Control-CLIP demonstrates better performance on both style and category recognition in specific domains. In addition, we propose an innovative cross-attention mechanism that utilizes the decoupled outputs of the style and category encoders to guide diffusion models. 
Compared to fine-tuning diffusion models, our approach preserves the general capabilities of pre-trained models while enhancing generation diversity and text fidelity. The plug-and-play design allows Control-CLIP to be easily and cost-effectively integrated into various text-to-image generation models.

\bibliographystyle{IEEEbib}
\bibliography{icme2025references}

\end{document}